\documentclass[11pt]{article}
\usepackage{geometry}
\usepackage{coling2020}
\usepackage{times}
\usepackage{url}
\usepackage{latexsym}
\usepackage{microtype}
\usepackage{url}
\usepackage{makecell}
\usepackage{hyperref}       
\usepackage{url}
\usepackage{booktabs}       
\usepackage{amsfonts}
\usepackage{nicefrac}      
\usepackage{microtype}   
\usepackage{lipsum}
\usepackage{graphicx}
\usepackage{verbatim}
\usepackage{multirow}
\usepackage{multicol}
\usepackage{color}
\usepackage{latexsym}
\usepackage{graphicx}
\usepackage{xcolor}
\usepackage{colortbl,booktabs}
\usepackage{tabu}
\usepackage{float}
\usepackage{amsmath}
\usepackage{diagbox}

\colingfinalcopy 

\title{Galileo at SemEval-2020 Task 12: Multi-lingual Learning for Offensive Language Identification using Pre-trained Language Models}

  \author{{Shuohuan Wang, Jiaxiang Liu, Xuan Ouyang, Yu Sun} \\
  Baidu Inc., China \\

  {\tt \{wangshuohuan,liujiaxiang,ouyangxuan,sunyu02\}@baidu.com } \\}

\date{}

\begin{document}
\maketitle
\begin{abstract}

 This paper describes Galileo’s performance in SemEval-2020 Task 12 on detecting and categorizing offensive language in social media. For Offensive Language Identification, we proposed a multi-lingual method using Pre-trained Language Models, ERNIE and XLM-R. For offensive language categorization, we proposed a knowledge distillation method trained on soft labels generated by several supervised models. Our team participated in all three sub-tasks. In Sub-task A - Offensive Language Identification, we ranked first in terms of average F1 scores in all languages. We are also the only team which ranked among the top three across all languages.  We also took the first place in Sub-task B - Automatic Categorization of Offense Types and Sub-task C - Offence Target Identification.
 
\end{abstract}

\blfootnote{
    \hspace{-0.65cm}  
    This work is licensed under a Creative Commons 
    Attribution 4.0 International License.
    License details:
    \url{http://creativecommons.org/licenses/by/4.0/}.
}

\section{Introduction}
Due to the growing number of Internet users, cyber-violence emerged with offensive language pervasive across social media. With anonymity as a “privilege”, netizens hide behind the screens, behaving in a manner most of them would not otherwise in reality. Thus, government organizations, online communities, and technology companies are all striving for ways to detect aggressive language in social media and help build a more friendly online environment.

Manual filtering is very time consuming and it can cause post-traumatic stress disorder-like symptoms to human annotators. One of the most common strategies \cite{waseem2017understanding,davidson2017automated,malmasi2018challenges,kumar2018benchmarking} to tackle the problem is to train systems capable of recognizing offensive content, which can then be deleted or set aside human moderation.

SemEval 2020 Task-12 \cite{zampieri-etal-2020-semeval} is the second edition of OffensEval \cite{zampieri2019semeval}. In this competition, organizers offers 5 languages datasets including Arabic \cite{mubarak2020arabic}, Danish \cite{sigurbergsson2020offensive}, English \cite{rosenthal2020}, Turkish \cite{coltekikin2020} and Greek \cite{pitenis2020}. In Sub-task A, the participants need to predict whether a post uses offensives language. Besides, the organizers provide other two sub-tasks which mainly focus on English, to predict the type and target of offensive language.

Participating in all 3 Sub-tasks, we proposed several methods based on pre-training language models including ERNIE and XLM-R. In Sub-task A, we scored 0.9199, 0.851, 0.8258, 0.802, 0.8989 in English, Greek, Turkish, Danish and Arabic respectively. We ranked first in average F1 scores, and ranked in top three across all languages. In Sub-task B and Sub-task C, we also took the first place with 0.7462 and 0.7145. In the following sections, we will elaborate the methods, dataset and experiments of our system.

\section{Relate Work}
\subsection{Monolingual Pre-trained Language Models}
Pre-training and fine-tuning have become a new paradigm in natural language processing, where the general knowledge is firstly learnt from large-scale corpus through self-supervised learning and then transferred to down-stream tasks for task-specific fine-tuning. The following are some representatives.
 
\cite{peters2018deep} proposed context-sensitive word vectors (ELMo) that enhance downstream tasks by acting as features. \cite{radford2018improving} proposed GPT which enhanced the context-sensitive embedding by adjusting the Transformer \cite{vaswani2017attention}. \cite{devlin2018bert} modeled a bidirectional language model (BERT) through a task similar to Cloze. \cite{yang2019xlnet} proposed a permuted language model (XLNet) which is a generalized autoregressive pre-training method. \cite{liu2019roberta} remove the next prediction task and pre-train longer to get a better pre-trained model (RoBERTa) . \cite{clark2020electra} proposed a method to joint generator and discriminator in ELECTRA. \cite{lan2019albert} and  \cite{raffel2019exploring} explored the larger model structure while optimizing the pre-training strategy in ALBERT and T5.

\cite{sun2019ernie} enhanced pre-trained language models with full masking of spans in ERNIE. \cite{sun2019ernie2} proposed continuous multi-task pre-training and several pre-training tasks in ERNIE 2.0. The researchers of ERNIE 2.0 released a new version recently which made a few improvements on knowledge masking and application-oriented tasks, with the aim to advance the model's general semantic representation capability. In order to improve the knowledge masking strategy, they proposed a new mutual information based dynamic knowledge masking algorithm. They also constructed pre-training tasks that are specific for different applications. For example, they added a coreference resolution task to identify all expressions in a text that refer to the same entity. For more details, please go to this website\footnote{http://research.baidu.com/Blog/index-view?id=128} .

\subsection{Cross-lingual Pre-trained Language Models}
In addition, there are also a lot of works on multilingual language models. \cite{devlin2018bert} provided a multilingual version of BERT that demonstrates surprising cross-language capabilities \cite{wu2019beto}. \cite{lample2019cross} proposed two tasks, Masked Language Model and Translation Language Model, to model monolingual corpus and bilingual parallel corpus respectively. \cite{huang2019unicoder} proposed Unicoder  incorporate more bilingual parallel corpus modeling methods.  \cite{song2019mass} and \cite{liu2020multilingual} proposed modeling methods that are more suitable for machine translation tasks in MASS and MBART. \cite{conneau2019unsupervised} used the ideas of RoBERTa in XLM-R and achieved better results than XLM.

\subsection{Methods of Offensive Language Detection and Categorization}

In the last few years, there have been several studies on the application of computational methods to cope with offensive language.
\cite{waseem2017understanding} proposed a typology that captures central  similarities and differences between subtasks. \cite{davidson2017automated} trained a multi-class classifier to distinguish between these different categories.
\cite{malmasi2018challenges} employed supervised classification along with a set of features that includes n-grams, skip-grams and clustering-based word representations. There are also several workshops for this problem. Such as   AWL\footnote{https://sites.google.com/view/alw3/} and TRAC\footnote{https://sites.google.com/view/trac1/home} \cite{kumar2018benchmarking} . 

Besides, there are several works focus on offensive language identification in languages other than English, such as Chinese \cite{su2017}, Dutch \cite{tulkens2016dictionary}, German \cite{ross2016measuring}, Slovene \cite{fiser2017} and Arabic \cite{mubarak2017}. There are also some researches \cite{basile2019semeval,mandl2019overview} about multilingual offensive language identification.

\section{Methodology}

\begin{figure}[h]
\centerline
{\includegraphics[width=11cm]{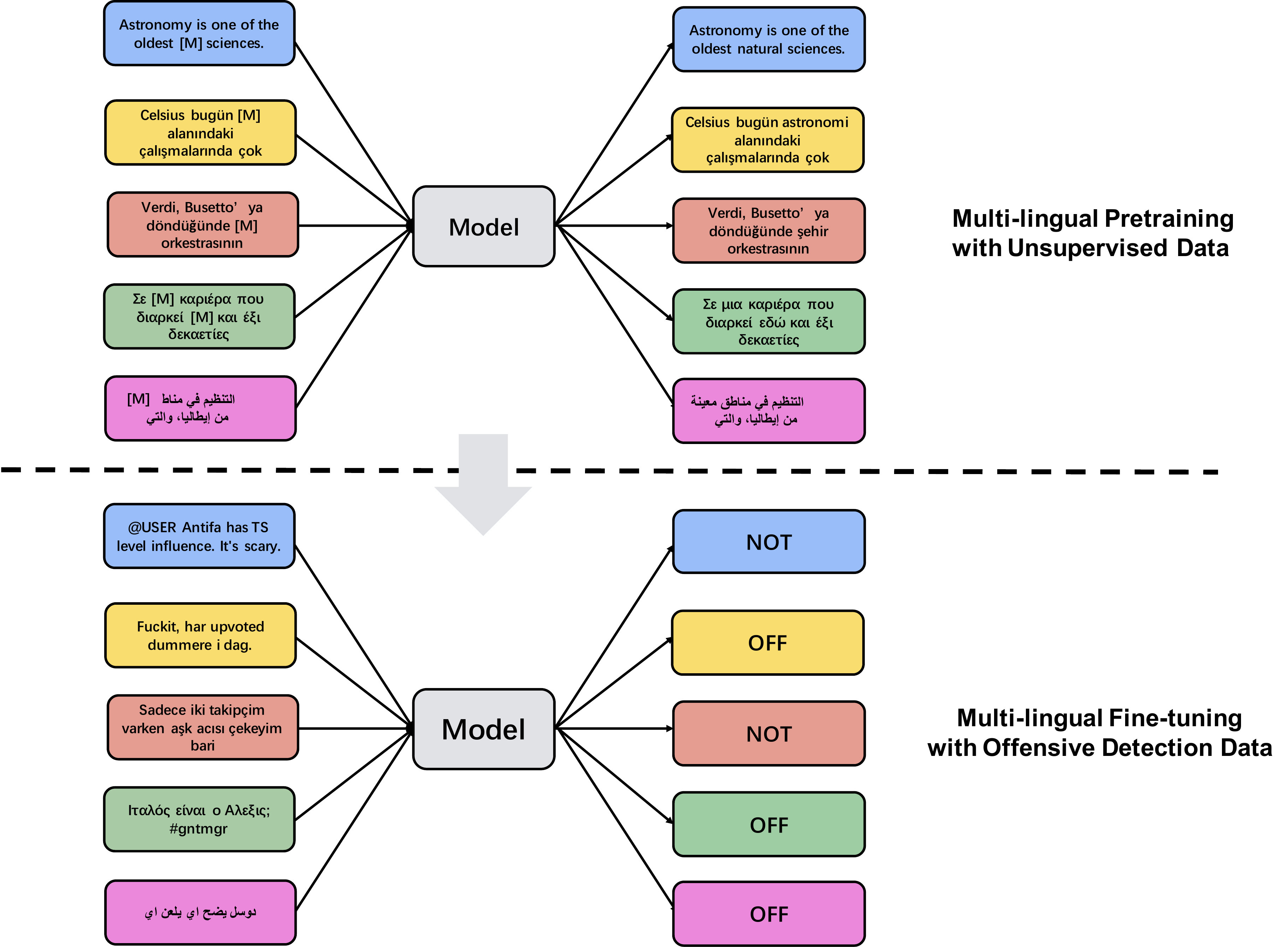}}
\caption{The Framework of Multi-lingual Offensive Language Detection}
\label{framework}
\end{figure}

\subsection{Multi-lingual Offensive Language Detection}

In Sub-task A, we expected to build a unified approach to detect offensive language in all languages.

Our algorithm has two steps. In the first step, pre-training using large scale multilingual unsupervised texts yields a unified pre-training model that can learn all the language representations together. In the second step, the pre-trained model was fine-tuned with labeled data. The detailed process is shown in Figure \ref{framework}.

To skip large-scale pre-training, we used an existing open-source model, XLM-R, as our first step. With Transformer as the backbone structure, XLM-R was pretrained with masked language model on Common Crawl dataset in over 100 languages. 

We add a full connected layer for classification upon the [CLS] position of the top layer of XLM-R, using the same parameter for all languages.

This approach can benefit from dataset in other languages and enhance the generality of the model. We will compare methods trained on multilingual data with those on monolingual data in Section 5.

\subsection{Offensive Language Categorization using Knowledge Distillation trained on Soft Labels}

\begin{figure} 
\centerline
{\includegraphics[width=11cm]{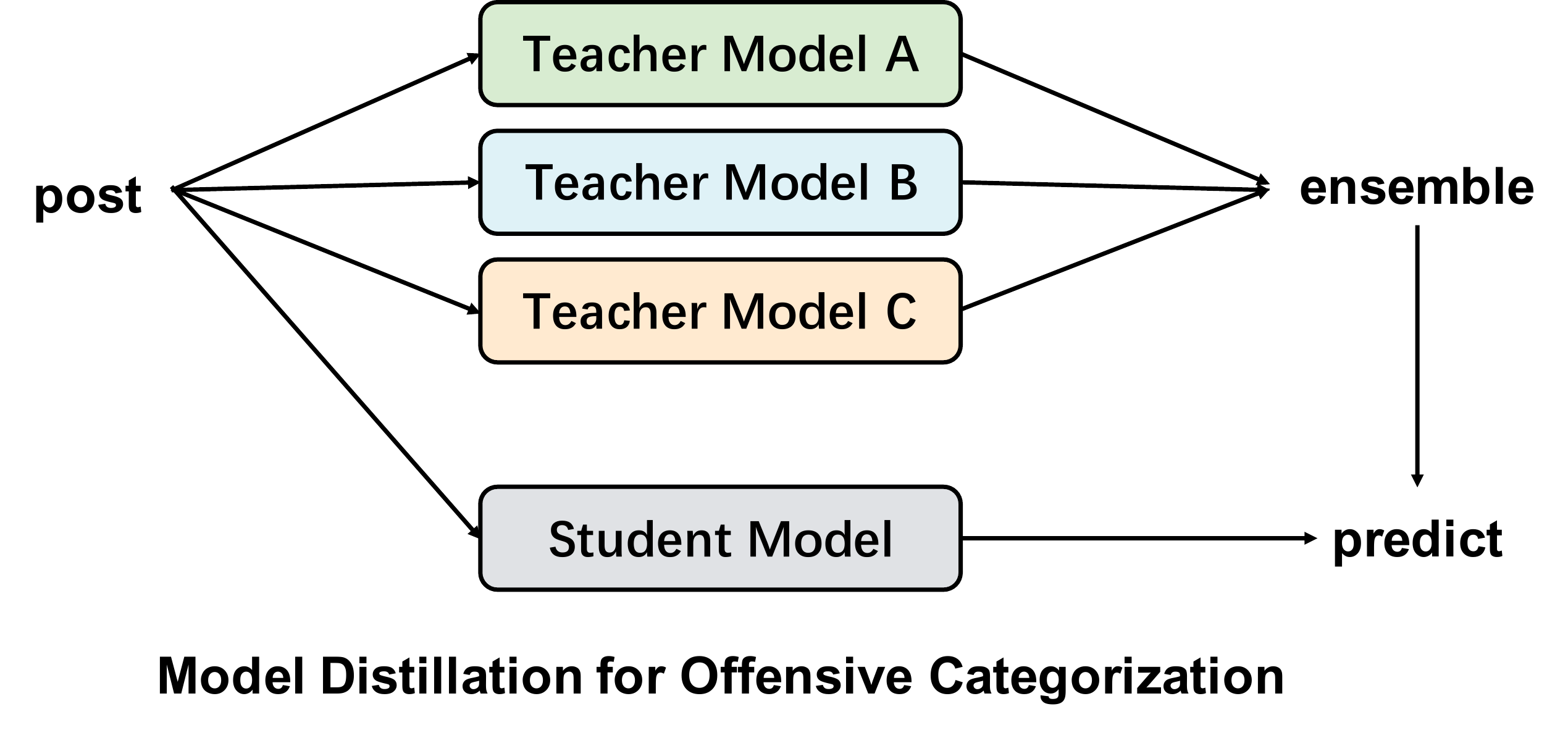}}
\caption{The Framework of Offensive Language Categorization}
\label{framework_2}
\end{figure}

In Sub-task B and Sub-task C, we constructed a knowledge distillation approach \cite{hinton2015distilling,davidson2017automated,liu2019improving} . Several supervised models provided calculated the probability of each label and generated a weighted probability (here we call it soft label). Then the student model was trained on those soft labels. Detailed process is shown in Figure \ref{framework_2}.

Suppose that $X$ is the contextual embedding of the token [CLS], which can be viewed as the semantic representation of input sentence.  Let $Q(c|X)$ be the class probabilities produced by the ensemble of several supervised models. The probability $P_r(c|X)$  that X is labeled as class c is predicted by a softmax layer. We use the standard cross entropy loss to learn the soft target:

\begin{equation}
Loss = -\sum_c Q(c|X) \log(P_r(c|X))
\label{eqn:cross-entropy-loss-soft}
\end{equation}

We used ERNIE 2.0 and ALBERT as our candidates of pre-training language models in Sub-task B and Sub-task C.

\section{Dataset}
We used datasets of OffensEval 2019 and OffensEval 2020 as our training data.
In OffensEval 2019, the organizers provide a dataset containing English tweets annotated using a hierarchical three-level annotation. 
In OffensEval 2020, the organizers did not provide additional data in English for training. They provided training data for four other languages, Turkish, Danish, Greek and Arabic. In addition, they provide a large amount of weakly labeled data generated by several supervised models.

\subsection{Sub-Task A - Offensive Language Identification}
In Sub-task A, the goal is to discriminate between offensive and non-offensive posts. Offensive posts include insults, threats, and posts containing any form of untargeted profanity. Each instance is assigned one of the following two labels. ’NOT’ means posts which do not contain offense or profanity. ’OFF’ means posts containing offense any form of non-acceptable language or a targeted offense.

In order to avoid uneven proportions of data across languages, we did not use the unannotated English data from OffensEval 2020. Instead, we used a mix of English data from OffensEval 2019 (including training data and test data) and training data from OffensEval 2020 in the other 4 languages as our training data. Details are shown in Table \ref{task_data_1}.

\begin{table}[h]
\begin{center}
\begin{tabular}{c||r|r|r||r|r|r}
\hline \bf \multirow{2}{*}{Languages} &  \multicolumn{3}{c||}{\bf{Train}} &  \multicolumn{3}{c}{\bf{Test}} \\ \cline{2-7}
 & \bf{OFF} & \bf{NOT} & \bf{TOTAL} & \bf{OFF} & \bf{NOT} & \bf{TOTAL}\\ \hline\hline
\bf{English} & 4640 & 9460 & 14100  & 1080 & 2807 & 3887 \\ \hline
\bf{Turkish} & 6131 & 25625 & 31756 & 716 & 2812 & 3528 \\ \hline
\bf{Arabic} & 1589 & 6411 & 8000 & 402 & 1598 & 2000 \\ \hline
\bf{Danish} & 384 & 2577 & 2961 & 41 & 288 & 329 \\ \hline
\bf{Greek} & 2486 & 6257 & 8743 & 242 & 1302 & 1546 \\ \hline
\end{tabular}
\end{center}
\caption{\label{task_data_1}  Dataset Statistics for Sub-task A }
\end{table}

\subsection{Sub-Task B - Automatic Offense Language Categorization }
In Sub-task B, the goal is to predict the type of offense. There are two types in sub-task B are the following. 'TIN' means posts containing an insult or threat to an individual, group, or others. 'UNT' means posts containing non-targeted profanity and swearing.  The dataset consists of two parts, a small portion of the manually annotated dataset from OffensEval 2019 and a large portion of the dataset from OffensEval 2020 constructed based on multiple supervision models. All the  training data in OffensEval 2020 provides the confidence that it has a target to attack. Details are shown in Table \ref{task_data_2}.

\begin{table}[h]
\begin{center}
\begin{tabular}{c||r|r|r||r|r|r}
\hline \bf \multirow{2}{*}{DataSet} &  \multicolumn{3}{|c||}{\bf{Train}} &  \multicolumn{3}{c}{\bf{Test}} \\ \cline{2-7}
 & \bf{TIN} & \bf{UNT} & \bf{TOTAL} & \bf{TIN} & \bf{UNT} & \bf{TOTAL}\\ \hline\hline
 \bf{OffensEval 2019} & \multirow{1}{*}{4089} & \multirow{1}{*}{551} & \multirow{1}{*}{4640} & \multirow{1}{*}{-} & \multirow{1}{*}{-} & \multirow{1}{*}{-} \\ \hline

 \bf{OffensEval 2020} & \multirow{1}{*}{149550} & \multirow{1}{*}{39424} & \multirow{1}{*}{188974} & \multirow{1}{*}{850} & \multirow{1}{*}{572} & \multirow{1}{*}{1422} \\ \hline

\end{tabular}
\end{center}
\caption{\label{task_data_2} Dataset Statistics for Sub-Task B  }
\end{table}

\subsection{Sub-Task C - Offense Target Identification}
In Sub-Task C, the goal is to predict the target of offense. The three labels in Sub-task C are the following. ’IND’ means posts targeting an individual. ’GRP’ means the target of these offensive posts is a group of people. ’OTH’ means the target of these offensive posts does not belong to any of the previous two categories. As with Sub-task B, all training data in OffensEval 2020 provide the confidence level for each label. Details are shown in Table \ref{task_data_c}.

\begin{table}[h]
\begin{center}
\begin{tabular}{c||r|r|r|r||r|r|r|r}
\hline \bf \multirow{2}{*}{DataSet} &  \multicolumn{4}{c||}{\bf{Train}} &  \multicolumn{4}{c}{\bf{Test}} \\ \cline{2-9}
 & \bf{IND} & \bf{GRP} & \bf{OTH} & \bf{TOTAL} & \bf{IND} & \bf{GRP} & \bf{OTH} & \bf{TOTAL}\\ \hline\hline
\bf{OffensEval 2019} & \multirow{1}{*}{2507} & \multirow{1}{*}{1152} & \multirow{1}{*}{430} & \multirow{1}{*}{4089} & \multirow{1}{*}{-} & \multirow{1}{*}{-} & \multirow{1}{*}{-}  & \multirow{1}{*}{-}\\ \hline

\bf{OffensEval 2020} & \multirow{1}{*}{152562} & \multirow{1}{*}{24917} & \multirow{1}{*}{11494} & \multirow{1}{*}{188973} &  \multirow{1}{*}{580} & \multirow{1}{*}{190}  & \multirow{1}{*}{80} & \multirow{1}{*}{850}\\ \hline

\end{tabular}
\end{center}
\caption{\label{task_data_c} Dataset Statistics for Sub-Task C }
\end{table}

\section{Experiments}
\subsection{Results of Sub-task A}
We validated our proposed methods based on two models, XLM-R Base and XLM-R Large. The metric used is the average F1 score of all labels. To make the results more reliable, we repeated the experiment 5 times and used the average F1 score. In Table \ref{result-taska}, we can see that the result of multilingual fine-tuning is better in all languages except Turkish. It might be caused by it taking up the largest proportion among all languages, leading to data of other languages being ignored. In the table below, we also listed the final submitted results and ranks in the contest, where we used ten-fold cross-validation-based ensemble of XLM-R Large.

\begin{table}[h]
\begin{center}
\begin{tabular}{c||c|c||c|c||c|c}
\hline \bf \multirow{2}{*}{Languages} &  \multicolumn{2}{c||}{\bf{XLM-R \small{BASE}}} &  \multicolumn{2}{c||}{\bf{XLM-R \small{LARGE}}} &  \multicolumn{2}{c}{\bf{Submitted Result(Ensemble)}}  \\ \cline{2-7}
 & \bf{Single (F1)} & \bf{Multi (F1)} & \bf{Single(F1)} & \bf{Multi(F1)} & \bf{Multi(F1)} & \bf{rank in all teams} \\ \hline\hline
\bf{English} & 0.9150 & 0.9214 & 0.9186 & \bf{0.9255} & 0.9199 & 3  \\ \hline
\bf{Turkish} & 0.8081 & 0.8084 & \bf{0.8265} & 0.8224 & 0.8258 & 1 \\ \hline
\bf{Arabic} & 0.8649 & 0.8730 & 0.8969 & \bf{0.9015} & 0.8989 & 3  \\ \hline
\bf{Danish} & 0.7733 & 0.7922 &0.7908 & \bf{0.8136} & 0.8020 & 2  \\ \hline
\bf{Greek} & 0.8266 & 0.8356 & 0.8356 & \bf{0.8392} & 0.8510 & 2\\ \hline\hline
\bf{Average} & 0.8376 & 0.8461 & 0.8537  & \bf{0.8604} & 0.8595 & 1 \\ \hline
\end{tabular}
\end{center}
\caption{\label{result-taska} Results for Sub-Task A }
\end{table}

\subsection{Results of Sub-task B and Sub-task C}
In both Sub-Task B and Sub-Task C, we made a comparison between hard target-based approach and soft target-based approach. Two models were used for validation, which are ALBERT-XXLarge and ERNIE 2.0. The results are shown in Table \ref{result-taskb} and Table \ref{result-taskc}, where it can be seen that the knowledge distillation approach is helpful for offensive categorization.

Same with Sub-Task A, the metric used is the average F1 score of all labels. Again, to make it more reliable, the average score of 5 repeated experiments was adopted. We also listed our final submitted results below, which were obtained using ten-fold cross-validation-based ensemble of ERNIE 2.0.

\begin{table}[h]
\begin{center}
\begin{tabular}{c|c|c}
\hline \bf \bf{Method} &  \bf{ALBERT(F1)} & \bf{ERNIE 2.0(F1)}  \\ \hline\hline
\bf{Learning with Hard Target } & 0.6810 & 0.6883 \\ \hline
\bf{Learning with Soft Target} & 0.7043 & \bf{0.7124}\\ \hline
\bf{Submitted Result} & - & 0.7462 \\ \hline

\end{tabular}
\end{center}
\caption{\label{result-taskb} Results for Sub-task B }
\end{table}

\begin{table}[h]
\begin{center} 
\begin{tabular}{c|c|c}
\hline \bf \bf{Method} &  \bf{ALBERT(F1)} & \bf{ERNIE 2.0(F1)}  \\ \hline\hline
\bf{Learning with Hard Target } & 0.6727 & 0.6773\\ \hline
\bf{Learning with Soft Target} & 0.6864 & \bf{0.6894} \\ \hline
\bf{Submitted Result} & - & 0.7145 \\ \hline

\end{tabular}
\end{center}
\caption{\label{result-taskc} Results for Sub-task C }
\end{table}

\label{intro}

\blfootnote{
    \hspace{-0.65cm}  
}

\section{Conclusion}
In this paper, we presented our approach on detecting and categorizing offensive language in social media. We proposed a multi-lingual learning method to detect offensive language and a knowledge distillation method to categorize offensive language. We will further our exploration of multilingual offensive language identification in future, e.g. validating the zero-shot performance of our model in more languages.

\bibliographystyle{coling}
\bibliography{semeval2020}

\end{document}